\documentclass[10pt,twocolumn,letterpaper]{article}

\usepackage{cvpr}
\usepackage{times}
\usepackage{epsfig}
\usepackage{graphicx}
\usepackage{amsmath}
\usepackage{amssymb}
\usepackage{multirow}
\usepackage{xcolor}
\usepackage{url}

\usepackage[breaklinks=true,bookmarks=false]{hyperref}

\cvprfinalcopy 

\def\cvprPaperID{71} 

\ifcvprfinal\fi
\begin{document}

\title{Hierarchical Back Projection Network for Image Super-Resolution}

\author{Zhi-Song Liu, Li-Wen Wang, Chu-Tak Li, and Wan-Chi Siu\\
The Hong Kong Polytechnic University\\
Hung Hom, Hong Kong\\
{\tt\small \{zhisong.liu, liwen.wang, ron.li\}@connect.polyu.hk, enwcsiu@polyu.edu.hk}
}

\maketitle
\thispagestyle{empty}

\begin{abstract}
   Deep learning based single image super-resolution methods use a large number of training datasets and have recently achieved great quality progress both quantitatively and qualitatively. Most deep networks focus on nonlinear mapping from low-resolution inputs to high-resolution outputs via residual learning without exploring the feature abstraction and analysis. We propose a Hierarchical Back Projection Network (HBPN), that cascades multiple HourGlass (HG) modules to bottom-up and top-down process features across all scales to capture various spatial correlations and then consolidates the best representation for reconstruction. We adopt the back projection blocks in our proposed network to provide the error correlated up- and down-sampling process to replace simple deconvolution and pooling process for better estimation. A new Softmax based Weighted Reconstruction (WR) process is used to combine the outputs of HG modules to further improve super-resolution. Experimental results on various datasets (including the validation dataset, NTIRE2019, of the Real Image Super-resolution Challenge) show that our proposed approach can achieve and improve the performance of the state-of-the-art methods for different scaling factors.
\end{abstract}

\section{Introduction}

Single Image Super-Resolution (SISR) attracts a lot of attention in the research community in the past few years. It is a fundamental low-level vision problem where the aim is to form a high-resolution (HR) image $\mathbf{Y}$ from a low-resolution (LR) image $\mathbf{X}$. Usually, SISR is described as an ill-posed problem $\mathbf{X}=\mathbf{H}\mathbf{Y}+\mu$, where $\mathbf{H}$ is a down-sampling operator, $\mu$ is additive white Gaussian noise with standard deviation $\sigma$. 

To resolve the ill-posed problem, Super-Resolution (SR) images can be obtained in the perspective of model-based optimization ~\cite{GMM,K-SVD,NCSR,BM3D,WNNM} and discriminative learning methods ~\cite{SRCNN,VDSR,DRRN,LapSRN,EDSR,DBPN}. The model-based optimization can be formulated as,
\begin{small}
\begin{equation}
\mathbf{\hat{Y}}=\underset{\mathbf{Y}}{\arg\min} \frac{1}{2}\Arrowvert\mathbf{X}-\mathbf{HY}\Arrowvert^2+\lambda\Omega(\mathbf{Y}) \tag{1}
\label{Equation 1}
\end{equation}
\end{small}{}
where $\lambda$ is the regularization factor that controls the significance of the regularization term $\Omega(\mathbf{Y})$. Though model-based optimization methods are flexible to handle different SR condition and noise, they are usually time-consuming and require various priors. 

On the contrary, discriminative approaches use external or internal paired LR-HR training samples to directly learn the nonlinear relationship. The objective is given by
\begin{small}
\begin{equation}
\underset{\Theta}{\min}\ell(\mathbf{\hat{Y}},\mathbf{Y}) s.t. \mathbf{\hat{Y}}=\underset{\mathbf{Y}}{\arg\min} \frac{1}{2}\Arrowvert\mathbf{Y}-\mathbf{WX}\Arrowvert^2+\lambda\Omega(\mathbf{W}) \tag{2}
\label{Equation 2}
\end{equation}
\end{small}
where $\mathbf{W}$ is the mapping model for reconstruction. The fidelity term $\underset{\mathbf{Y}}{\arg\min} \frac{1}{2}\Arrowvert\mathbf{Y}-\mathbf{WX}\Arrowvert^2$ determines the distortion of reconstruction and similarly, the regularization term $\Omega(\mathbf{W})$ controls the complexity of the mapping model. In the previous research works, patch-based approaches use classification tools , like kNN ~\cite{Chang2004}, to classify the patches from natural images and capture the mapping relationship for clustered patches. Taking the advantage of non-local statistical priors from external datasets, there are many successful approaches that achieve good SR performance by off-line training classifiers and regressors for efficient on-line reconstruction. For example, Timofte \etal ~\cite{Timofte2013,Timofte2015} proposed the adjusted anchored neighbor regression (ANR and A+) which uses clustering on encoded sparse dictionary to search nearest neighbor dictionary atoms for LR patch reconstruction. Siu \etal ~\cite{Junjie15,Junjie17,ISCAS17,ICIP18} proposed random forests for binary classification to obtain fast and high qualified image SR and hierarchical decision trees to further boost up image SR performance. 

Since Dong \etal ~\cite{SRCNN} proposed the first deep convolution neural network (CNN) for image SR, a large number of CNN based SR approaches have been proposed to significantly improve the image SR performance. Along with the development of other computing vision fields, i.e., image classification, object detection and so on, more deep and complex models are adopted in image SR. For example, VDSR ~\cite{VDSR} uses a 20-layer convolution network for different up-sampling factors. Tai \etal ~\cite{DRRN} proposed a deep recursive residual network by using recursive blocks to explore long-term correlations between LR and HR images. LapSRN ~\cite{LapSRN} uses Laplacian pyramid networks to gradually super-resolve LR with different up-sampling factors. Most recently, Haris \etal ~\cite{DBPN} proposed Deep Back Projection Network (DBPN) for image SR by iteratively computing reconstruction errors then, fusing them back for model tuning. 

\begin{figure*}[h]
\vskip 0.01in
\begin{center}
\centerline{\includegraphics[width=0.6\textwidth]{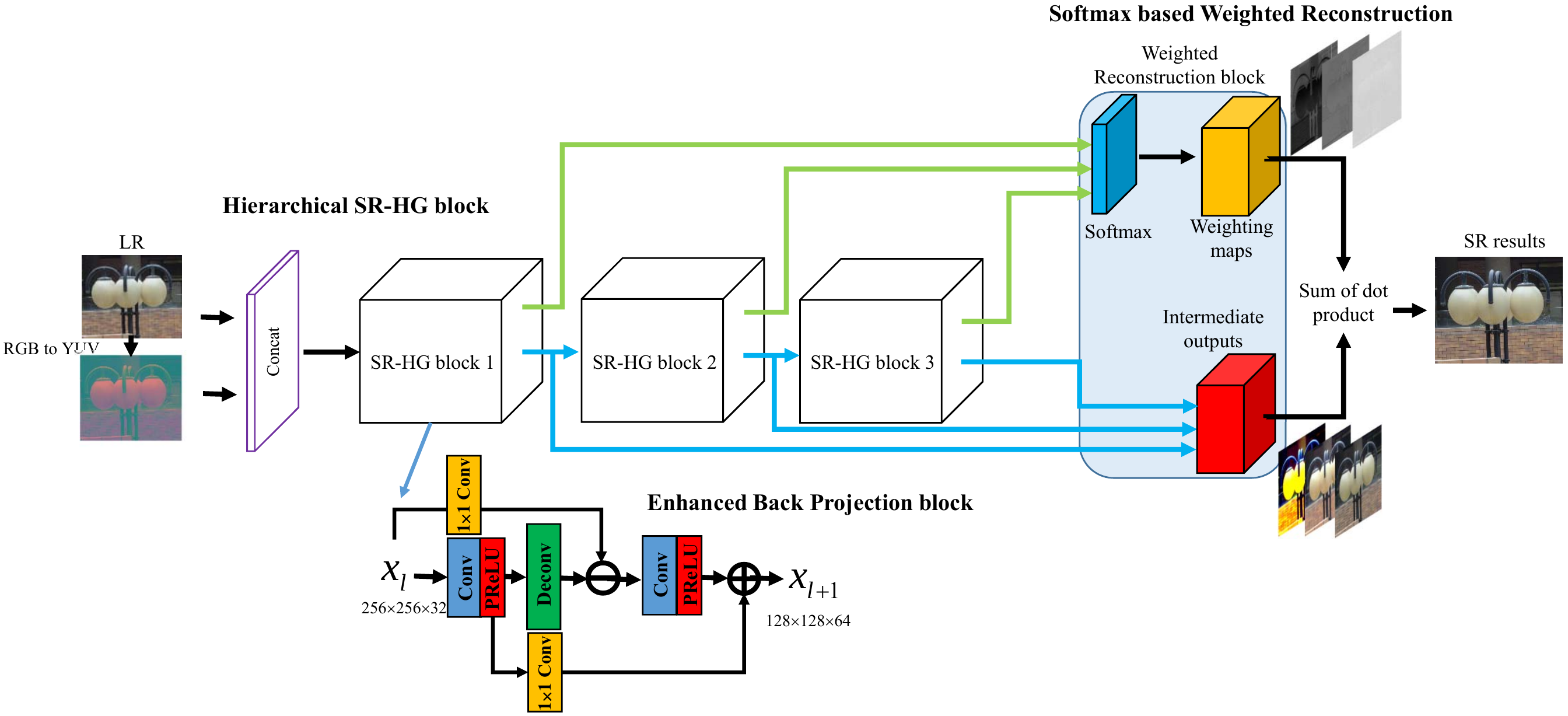}}
\caption{Proposed HBPN structure. In a bottom-up and top-down manner, it can explore various scales to extract hierarchical features for image SR.}
\label{Figure 1}
\end{center}
\vskip -0.3in
\end{figure*}

Inspired by~\cite{DBPN}, we design a Hierarchical Back Projection Network (HBPN) for image SR. As shown in Figure~\ref{Figure 1}, our work have following contributions:
\begin{itemize}
\item \textbf{Enhanced back projection blocks}. We propose an enhanced back projection block, including the new Up-sampling Back Projection block (UBP) and Down-sampling Back Projection block (DBP). Both UBP and DBP embed the back projection mechanism in the residual block to update up-sampling and down-sampling errors for better results. The key modification is two 1$\times$1 convolution layers within the back projection block to fine tune the LR and HR features. Details are explained in Section 3.
\item \textbf{Hierarchical SR HourGlass (SR-HG) module}. We stack multiple stages of SR-HG modules to capture various spatial correlations by repeated bottom-up and top-down process across all scales. Different from HG structure used in other applications ~\cite{HG1,HG2}, we replace the pooling and deconvolution layer by enhanced back projection blocks for better feature down- and up-sampling process.
\item \textbf{Softmax based Weighted Reconstruction (WR)}. To encourage different SR-HG modules super-resolve LR images in a hierarchical order, each SR-HG module outputs one coarse SR result and one weighting map. At the final WR stage, we propose to use Softmax layer to normalize weighting maps from different SR-HG modules to obtain the global weighting map. Finally, we consolidate all coarse SR results by using the global weighting map to output the final SR image.
\end{itemize}

\section{Related Work}

In order to compare the different SR reconstruction measurements, we can divide the convolutional neural network based SR approaches into distortion based SR and perception based SR.

\subsection{Distortion based image super-resolution}

As discussed in Section 1, to resolve Equation~\ref{Equation 2}, the end-to-end CNN model is a very direct and efficient method. By inputting LR images, we can define a mean squared errors based loss function to target on optimizing the convolutional parameters to obtain the SR outputs with minimal distortion. Considering the mismatch of dimension between LR and HR images, there are different designs of CNN models for SR. In the early stage of CNN for image SR, researchers inherited the knowledge on traditional machine learning based SR approaches by initially up-sampling LR images to the desired size by simple interpolation, i.e., Bicubic, and then learn the mapping model between the up-sampled LR and HR images. SRCNN ~\cite{SRCNN} and many other CNN approaches ~\cite{VDSR,DRRN,LapSRN,EDSR,DBPN} use this idea to build networks using cascaded convolution process. In order to grasp long-term correlation of pixels for reconstruction, we need to stack more convolution layers to cover a larger receptive field. However, building deeper convolution networks can encounter computation exploding and gradient vanishing problems. To resolve the former problem, Kim \etal ~\cite{DRCN} and Tai \etal ~\cite{DRRN} proposed to use recursive convolution networks to increase recursion depth rather than convolution depth without introducing new parameter for computation. For the latter problem, residual learning ~\cite{ResNet} is introduced in CNN models to add shortcuts to avoid gradient vanishing. In recent SR works, Lim \etal ~\cite{EDSR} proposed a state-of-the-art CNN network using residual blocks to achieve good SR performance on various datasets. 

Rather than using initial interpolation to up-sample LR image to feed into CNN for training, there have also been some novel CNN works that build up-sampling process into CNN models. The deconvolution with stride larger than 1 is used in CNN working as an up-sampling process, Lai \etal ~\cite{LapSRN} proposed a Laplacian Pyramid network to gradually super-resolve LR image by different scales. Shi \etal ~\cite{ESPCN}, on the other hand, proposed the sub-pixel convolution process to work as a pixel based interpolation for enlargement. Recently, Haris \etal ~\cite{DBPN} further studied the residual learning on image SR and proposed the back projection based residual block that can efficiently learn LR and HR feature maps iteratively to feedback residual errors. 

\subsection{Perception based image super-resolution}
Rather than targeting on minimizing mean squared errors based loss function, perception based image SR focuses on visual quality over data fidelity. Since a pioneer work on using the Generative Adversarial Network (GAN) for image SR ~\cite{SRGAN}, there are a lot of studies on using adversarial loss as a measurement for SR performance. By replacing the \textit{l}n-norm minimization by distribution divergence, we force the SR networks to learn the meaningful features rather than pixel differences. The idea of using GAN for image SR can be described as: the generator and discriminator learn from each other to generate a ``fake'' SR image that gives minimal distance on the high-level feature space (features used commonly extracted from VGG19 ~\cite{VGG}). Wang \etal ~\cite{ESRGAN} further investigated this study. They modified the generator by using Residual-in-Residual Dense Block to improve SR performance in terms of PSNR (one measurement of distortion) and then they fine tuned the network by using adversarial loss to generate SR image with better visual quality.

From recent studies of GAN for image SR, one of the key issues is still the design of generators. A good generator should be able to extract rich feature maps for estimation by any criteria. Our proposed network can also be considered as a perception based image SR by using adversarial loss. However, the measurement of visual quality was only used in 4$\times$ image SR ~\cite{SRGAN,EnhanceNet,ESRGAN}. To make a good comparison, we still use distortion based evaluation (PSNR, SSIM, etc.) to make analysis among different approaches.

\section{Hierarchical Back Projection Network}
Before introducing our proposed work, let us first define some terms. As defined in Section 1, given a RGB LR image $\mathbf{X}$ with size $h\times w\time 3$, we want to super-resolve it by $\alpha\times$ to the dimension $\alpha h\times \alpha w\time 3$, the HR image $\mathbf{Y}$. The super-resolved image is the SR image $\mathbf{\hat{Y}}$.

\subsection{Back projection}

\begin{figure}[t]
\vskip 0.01in
\begin{center}
\centerline{\includegraphics[width=0.9\columnwidth]{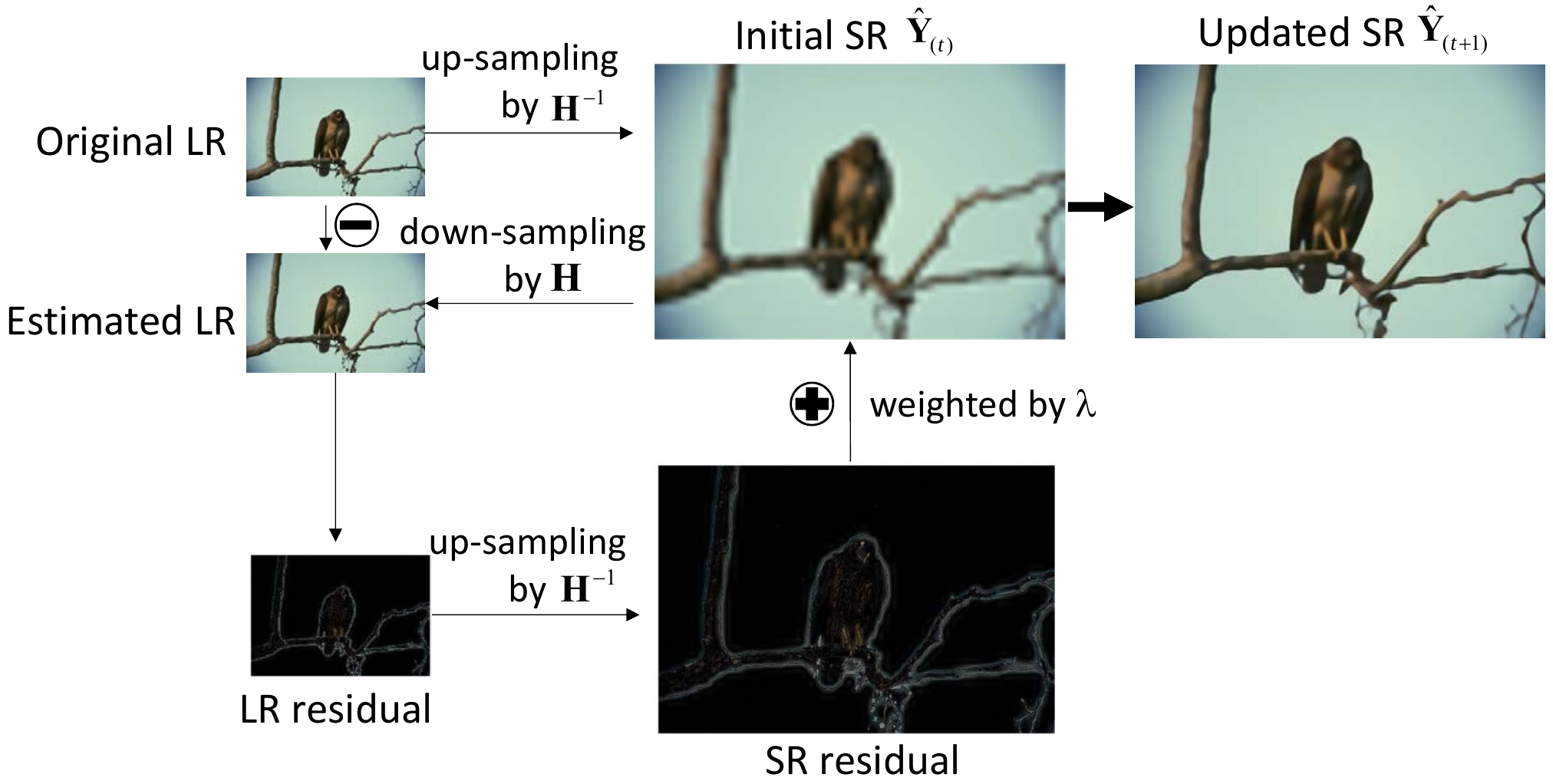}}
\caption{Back Projection procedure.}
\label{Figure 2}
\end{center}
\vskip -0.3in
\end{figure}

Let us first revisit the back projection approach that has been commonly used in image SR. Back projection was first proposed to utilize multiple LR images to estimate one SR image. ~\cite{DBPN} comes up with using back projection to refine SR image to improve the quality. It is an efficient iterative process to improve the data fidelity of SR by minimizing the loss between the original LR image and the down-sampled SR image. Mathematically, description of the back projection is
\begin{small}
\begin{equation}
\mathbf{\hat{Y}}_{\mathit{t}+1}=\mathbf{\hat{Y}}_{\mathit{t}}-\lambda\mathbf{H^{-1}}(\mathbf{H}\mathbf{\hat{Y}}_{\mathit{t}}-\mathbf{X}) \tag{3}
\label{Equation 3}
\end{equation}
\end{small}
where $\mathbf{H^{-1}}$ is the inverse operator of $\mathbf{H}$ which represents the up-sampling operation process. For estimating the SR residues, we need to assume a certain known down-sampling and up-sampling operators. $\lambda$ is the trade-off parameter to control the ratio of the residual information to gradually improve the SR quality. $\mathit{t}$ is the iteration number. A simple back projection process is shown in Figure~\ref{Figure 2}.
Back projection has been widely used in many SR approaches as a final refinement to reduce the distortion in terms of PSNR. However, it is observed that the down- and up-sampling operators need to be pre-determined as fixed parameters for estimation which may not obtain optimal results. To resolve this problem, ~\cite{DBPN} proposes to embed the back projection into CNN model to learn the unknown parameters by training. By using multiple proposed back projection blocks, it can expand the iterative process as a cascading process using more parameters to minimize the SR residual information. Our study further develops this work by coming up with a hierarchical back projection network to learn LR and HR features across different scales to extract more compact and robust features for reconstruction.

\subsection{Enhanced back projection blocks}
Let us propose our Enhanced back projection blocks, which contain both new Up-sampling Back Projection (UBP) and Down-sampling Back Projection (DBP) blocks. The UBP is the forward back projection process that estimates HR residues while the UBP is the backward back projection process that estimates LR residues. The details of two blocks are shown in Figure 3. 

\begin{figure}[htb]
\vskip 0.01in
\begin{center}
\centerline{\includegraphics[width=\columnwidth]{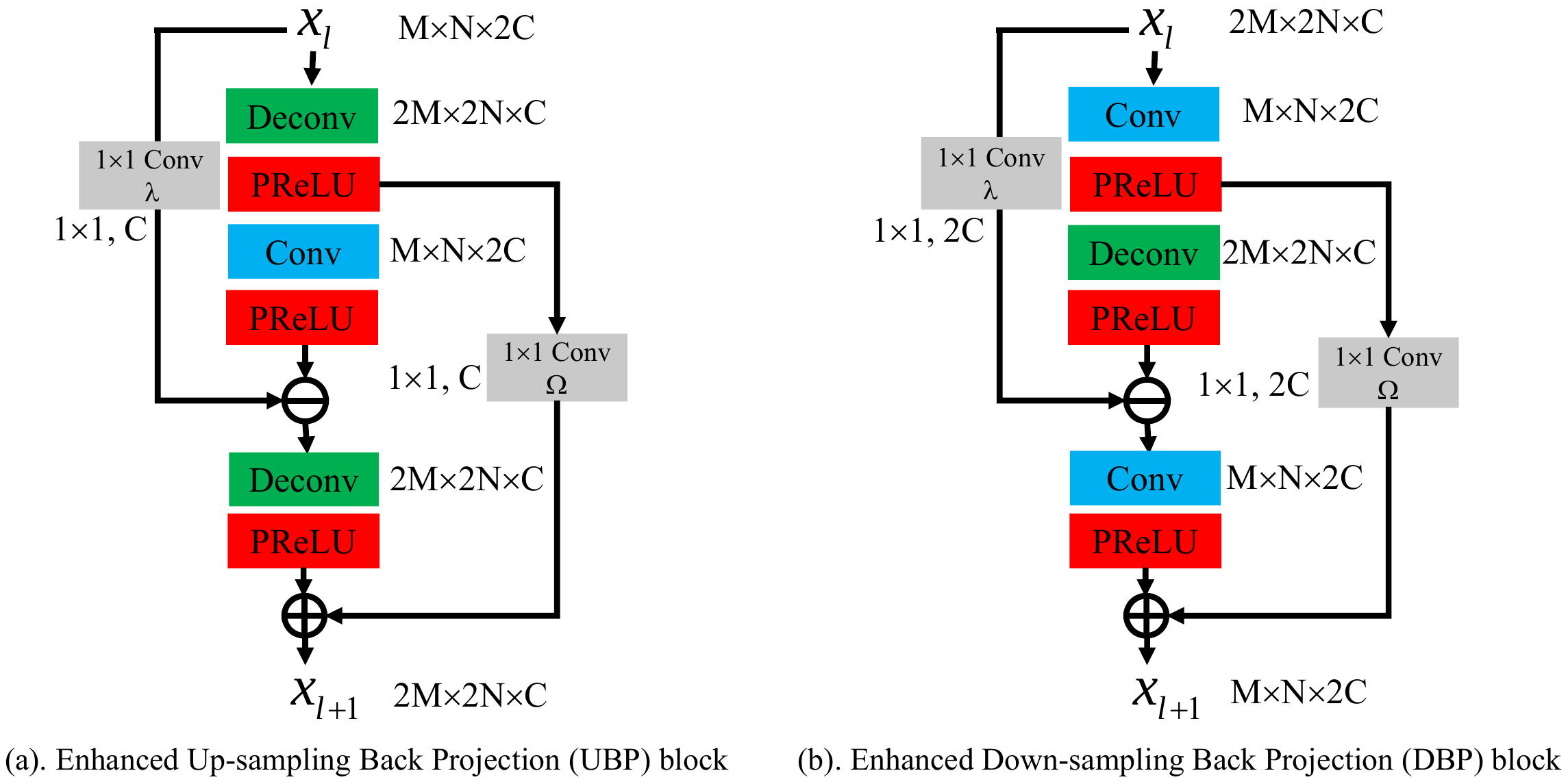}}
\caption{Proposed Enhanced back projection blocks: (a) the UBP block that up-samples the LR feature maps by 2$\times$ and reduces the number of feature maps by half, and (b) the DBP block that down-samples the HR feature maps by half and increases the number of feature maps by 2$\times$.}
\label{Figure 3}
\end{center}
\vskip -0.3in
\end{figure}

The process of UBP block can be described by rewriting Equation~~\ref{Equation 3} as,
\begin{small}
\begin{equation}
\mathit{x}_{\mathit{l}+1}=\Omega{\mathbf{D}\mathit{x_l}}+\mathbf{D}\left(\lambda\mathit{x_l}-\mathbf{CD}\mathit{x_l}\right) \tag{4}
\label{Equation 4}
\end{equation}
\end{small}
Similarly, the process of DBP block can be considered as the backward of UBP that estimates the LR residues as Equation~~\ref{Equation 5},
\begin{small}
\begin{equation}
\mathit{x}_{\mathit{l}+1}=\Omega{\mathbf{C}\mathit{x_l}}+\mathbf{C}\left(\lambda\mathit{x_l}-\mathbf{DC}\mathit{x_l}\right) \tag{5}
\label{Equation 5}
\end{equation}
\end{small}
There are two key modifications between our proposed Enhanced back projection blocks and that in DBPN~\cite{DBPN}: global weighting model $\Omega$ and residual weighting model $\lambda$. They all use $1\times1$ convolution layers to work as the weighting process. 

For the residual weighting model $\lambda$, it resembles the trade-off parameter in Equation~~\ref{Equation 3} that provides the regularization on the update of SR residues. Without followed by any activation function, this $1\times1$ convolution layer is a linear weighted model that can tune the residual information without increasing any computation burden.

For the global weighting model, it has two jobs: first, to work as a weighted model to tune the down- and up-sampled features for update so that we can introduce one extra freedom of parameters for training; second, to adjust the channel (number) of feature maps for addition. For example, from (a) in Figure~\ref{Figure 3}, the global weighting model reduces the number of feature maps by half. From (b) in Figure~\ref{Figure 3}, the global weighting model doubles the number of feature maps for addition. 

\subsection{Hierarchical SR HourGlass (SR-HG) module}
For the proposed SR-HG module, we adopt the HourGlass structure to cascade multiple enhanced back projection blocks in bottom-up and top-down manner. The HourGlass structure is commonly used in many computing vision fields. By down-sampling the size of feature maps while increasing the number of feature maps, we can extract denser and deeper features for various applications. The key differences of our proposed SR-HG module are three folds: 1) replacing pooling process by DBP blocks to avoid information loss, 2) replacing the single convolution process by DBP blocks to down-sample the feature maps and 3) output a coarse SR image and a weighting map.

\begin{figure}[htb]
\vskip 0.01in
\begin{center}
\centerline{\includegraphics[width=0.7\columnwidth]{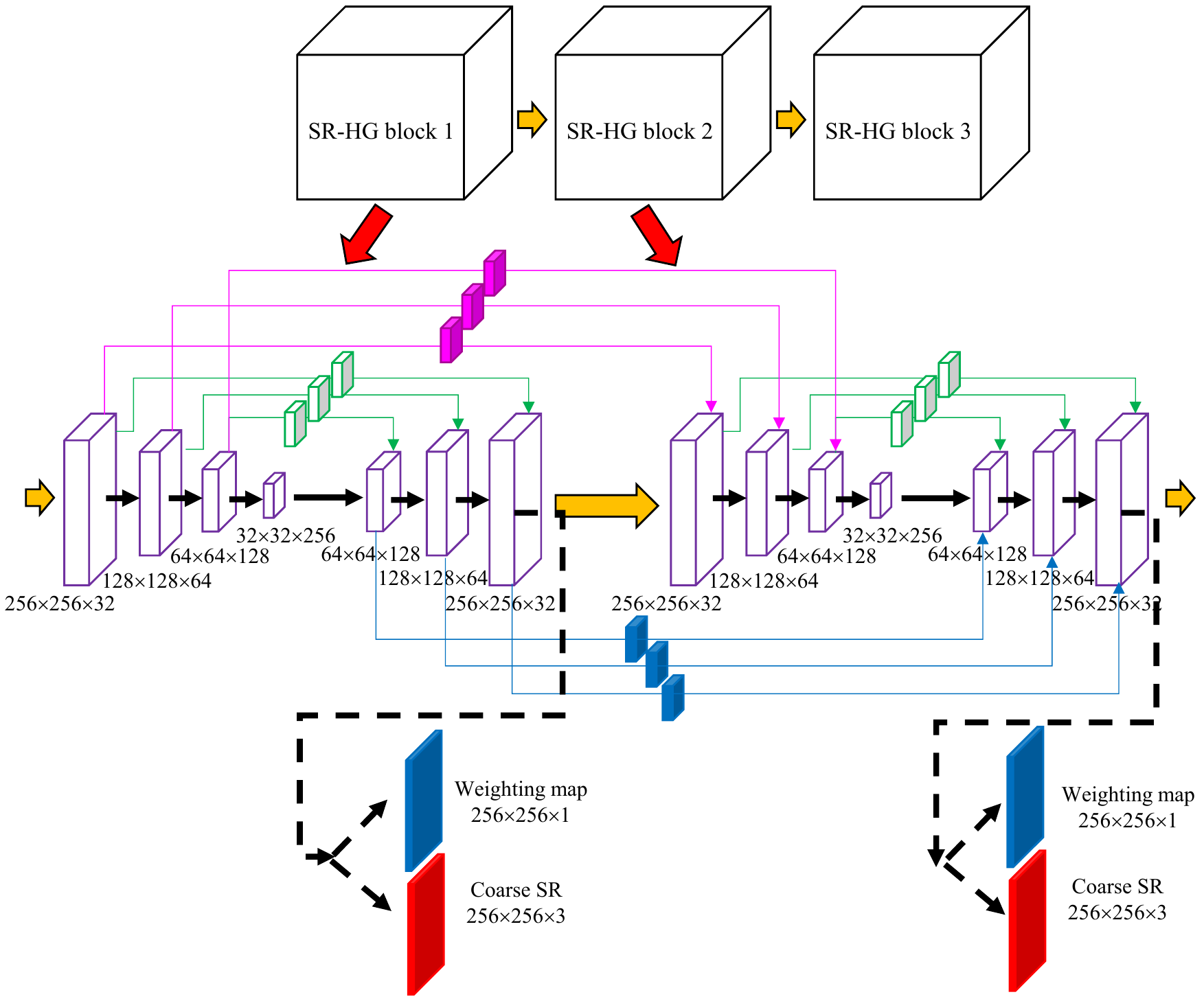}}
\caption{Proposed SR-HG module structure.}
\label{Figure 4}
\end{center}
\vskip -0.3in
\end{figure}

The complete structure of SR-HG is shown in Figure ~\ref{Figure 4}. For each SR-HG module, it contains 3 DBP blocks for down-sampling process and 3 UBP blocks for up-sampling process. For DBP and UBP blocks with same feature dimension, we use $1\times1$ convolution as local shortcuts (green blocks) to share the features. For different SR-HG blocks, we use $1\times1$ convolution as global shortcuts (pink and blue blocks) to share features across different modules. For each SR-HG module, there are two branches (dash lines in Figure~\ref{Figure 4}) to generate one coarse SR result and one weighting map to describe the contribution of the coarse SR. There are global and local shortcuts that share the features across different HourGlass modules and spatial scales. Each SR-HG module contains 3 UBP blocks for up-sampling and 3 DBP blocks for down-sampling and each UBP/DBP block up-/down-samples the input data by $2\times$. Totally, the input data are first down-sampled by $8\times$ and then up-sampled by $8\times$. In the meantime, the number of features are first increased by $8\times$ and then decreased by $8\times$ so that the network can learn denser and more compact features for reconstruction.

\subsection{Softmax based Weighted Reconstruction (WR)}
\begin{figure}[htb]
\vskip 0.01in
\begin{center}
\centerline{\includegraphics[width=0.8\columnwidth]{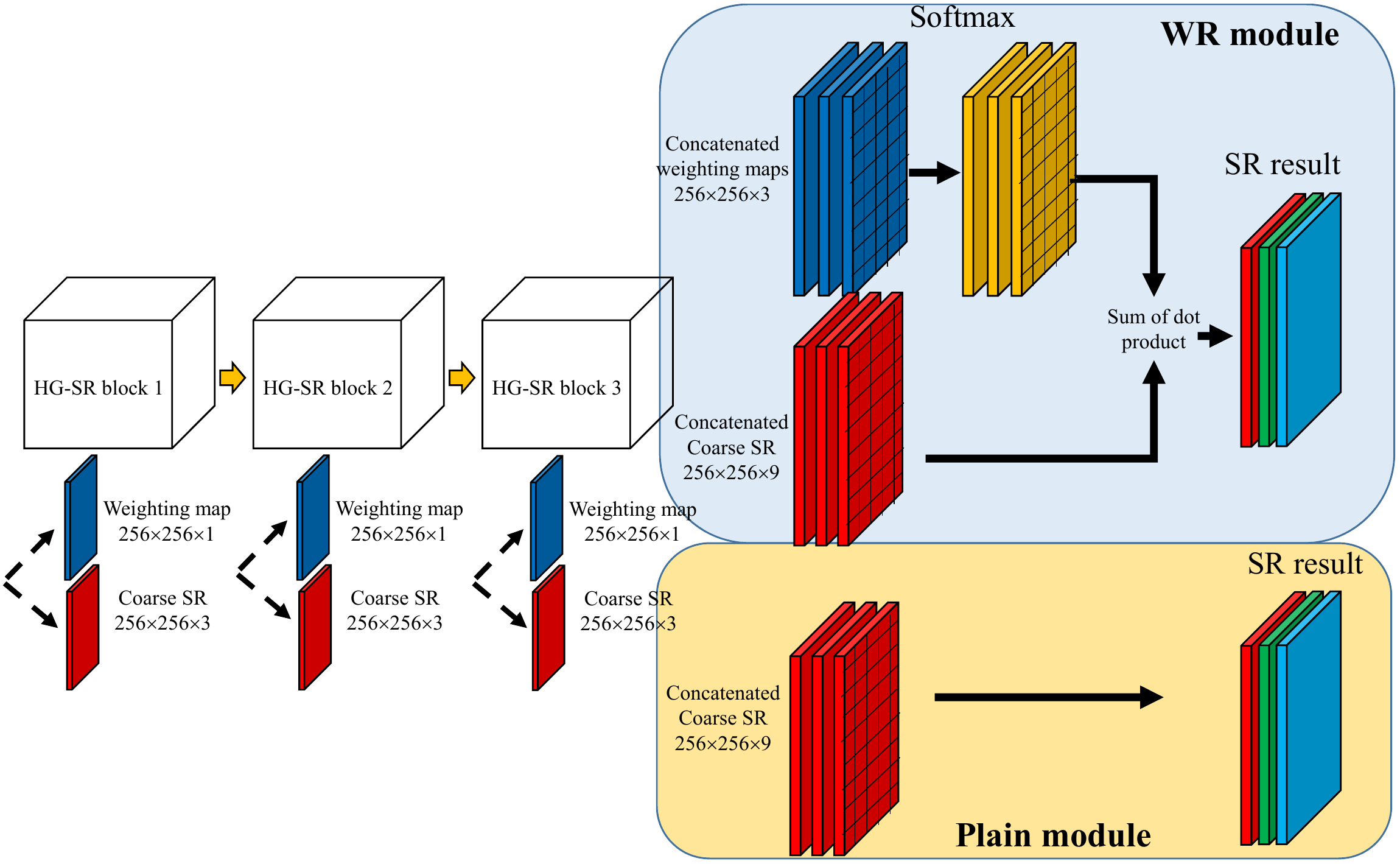}}
\caption{Proposed Weighted Reconstruction module.}
\label{Figure 5}
\end{center}
\vskip -0.3in
\end{figure}

For the final reconstruction, instead of concatenating coarse SR results from different SR-HG modules to generate the final SR by one convolution layer, we propose a Softmax based Weighted Reconstruction (WR) that makes use of the weighting maps to estimate the contribution of coarse SR results. It can be regarded as an adaptive weighted addition of coarse SR results. The comparison between WR process and plain process is shown in Figure~\ref{Figure 5}. It concatenates the weighting maps from SR-HG modules and learns a global probability map using a Softmax normalization. The coarse SR results are weighted by the probability map to generate the final SR image.

For the plain process, it simply concatenates the coarse SR results together and learns one convolution layer to output the SR results without considering the internal correlation between coarse SR results. In the WR module, the Softmax layer is used to normalize the weighting maps from SR-HG modules in the range of [0, 1]. Then the final SR image is the weighted sum of the coarse SR results. By using Softmax normalization, we force each SR-HG module to learn the SR image at different scales. 

\section{Experimental Results}

\subsection{Implementation and training setups}

Different from DBPN ~\cite{DBPN} which has different structures and configurations for different up-sampling enlargement, the proposed HBPN network uses the same structure as shown in Figure ~\ref{Figure 1}. In UBP and DBP blocks, we use $6\times6$ convolution filters with two striding and two padding for down- and up-sampling. For shortcut connections, we use $3\times3$ convolution filters with one striding and 1 padding. We initialize the weights based on ~\cite{PReLU}. The testing data include Set5 ~\cite{Set5}, Set14 ~\cite{Set14}, BSD100 ~\cite{BSD100}, Urban100 ~\cite{Urban100} and Manga109 ~\cite{Manga109} on $2\times$, $4\times$ and $8\times$ SR enlargement.

The training data include 800 2K images from DIV2K ~\cite{NTIRE} and 2650 2K images from Flickr ~\cite{EDSR}. Each image was rotated and flipped for augmentation to increase the images by 8$\times$. The LR images were down-sampled and initially up-sampled by \textit{bicubic} function in MATLAB on different scaling factors. We extracted LR-HR patch pairs from images of size $256\times256$. In order to achieve better SR performance, for different SR scaling factors, we trained our model by using different LR-HR training patches. The learning rate is set to 0.0001 for all layers. The batch size is 8 for every 5$\times10^5$ iterations and 32 for the rest 5$\times10^5$ iterations to achieve better results. For optimization, we used Adam with the momentum to 0.9 and the weight decay of 0.0001. All experiments were conducted using Caffe, MATLAB R2016b on two NVIDIA GTX1080Ti GPUs.


\subsection{Model analysis}
\textbf{Scaling factors of UBP and DBP.} For each SR-HG module, we used DBP blocks to down-sample the feature maps to the smallest size and UBP blocks as mirror reflection to up-sample feature maps to the original size. For input data with size $\mathit{M}\times \mathit{N}\times 3$, we used \textit{T} DBP blocks to down-sample the input to obtain feature maps with size $\frac{\mathit{M}}{2^{\mathit{T}}}\times \frac{\mathit{N}}{2^{\mathit{T}}}\times (64\cdot2^{\mathit{T}-1})$. To demonstrate the capability of this bottom-up and top-down structure, we conducted multiple networks \textit{HG}-1, \textit{HG}-2, \textit{HG}-3 (which is the proposed HBPN model) and \textit{HG}-4 for 4$\times$ enlargement on Set5 to make comparison.

\begin{figure}[h]
\vskip 0.01in
\begin{center}
\centerline{\includegraphics[width=0.7\columnwidth]{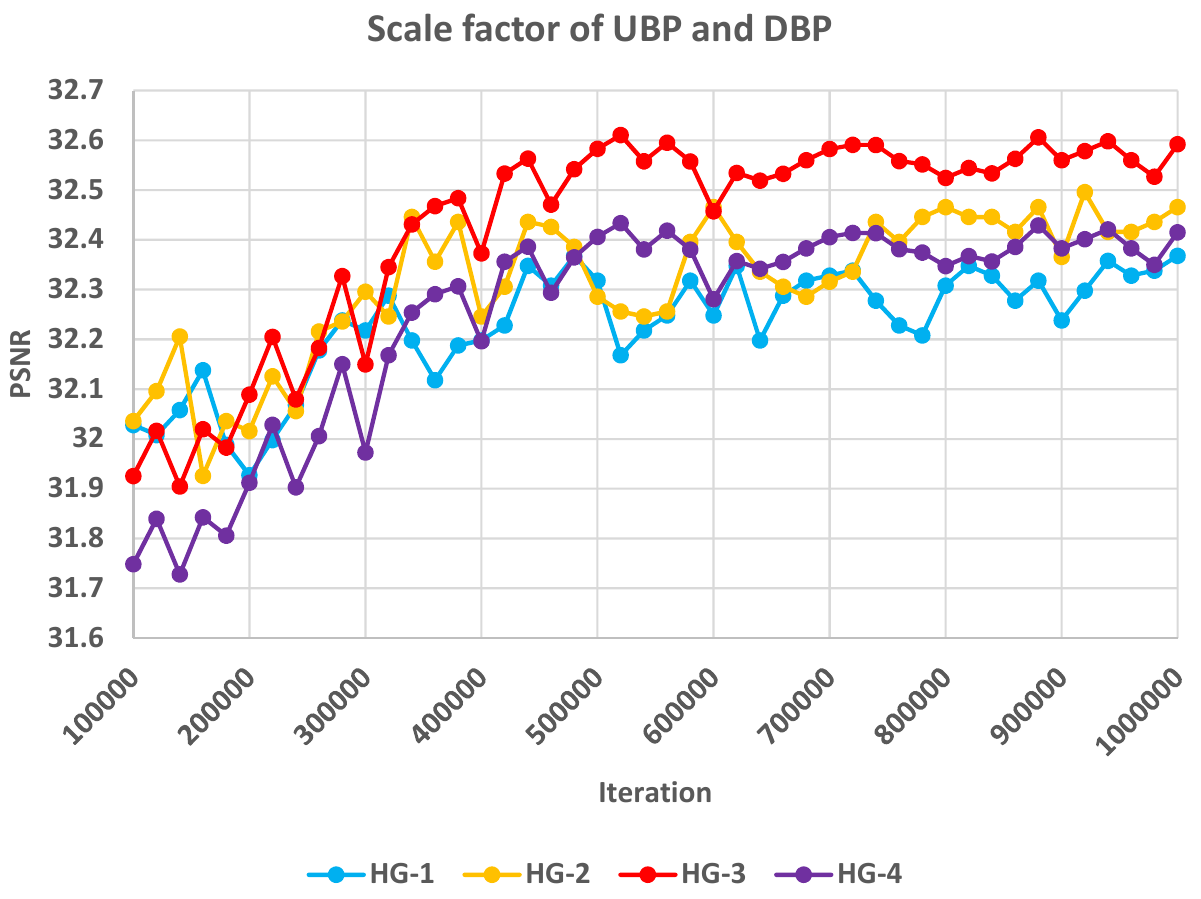}}
\caption{Back projection blocks analysis with different networks.}
\label{Figure 6}
\end{center}
\vskip -0.3in
\end{figure}

The results are shown in Figure~\ref{Figure 6}. We compare different SR-HG blocks using different numbers of UBP and DBP to down- and up-scale features. Using \textit{HG}-3 shows the best performance comparing with other networks. Due to the model complexity, \textit{HG}-1 and \textit{HG}-2 can converge faster than \textit{HG}-3 and \textit{HG}-4. As the best performance, \textit{HG}-3 achieves 32.66 dB in terms of PSNR which is 0.2 dB and 0.4 dB better than \textit{HG}-2 and \textit{HG}-4. 

\textbf{Number of SR-HG modules.} Generally, a deeper network can train more parameters to learn deeper feature representation for good performance. By stacking more and more SR-HG modules, ~\cite{DBPN} shows that the network with more HG blocks can produce a better prediction. In our experiments, we conduct multiple networks with different number of SR-HG modules: \textit{S} (2 SR-HG modules), \textit{M} (3 SR-HG modules, which is the proposed HBPN model) and \textit{L} (4 SR-HG modules). 

\begin{figure}[t]
\vskip 0.01in
\begin{center}
\centerline{\includegraphics[width=0.7\columnwidth]{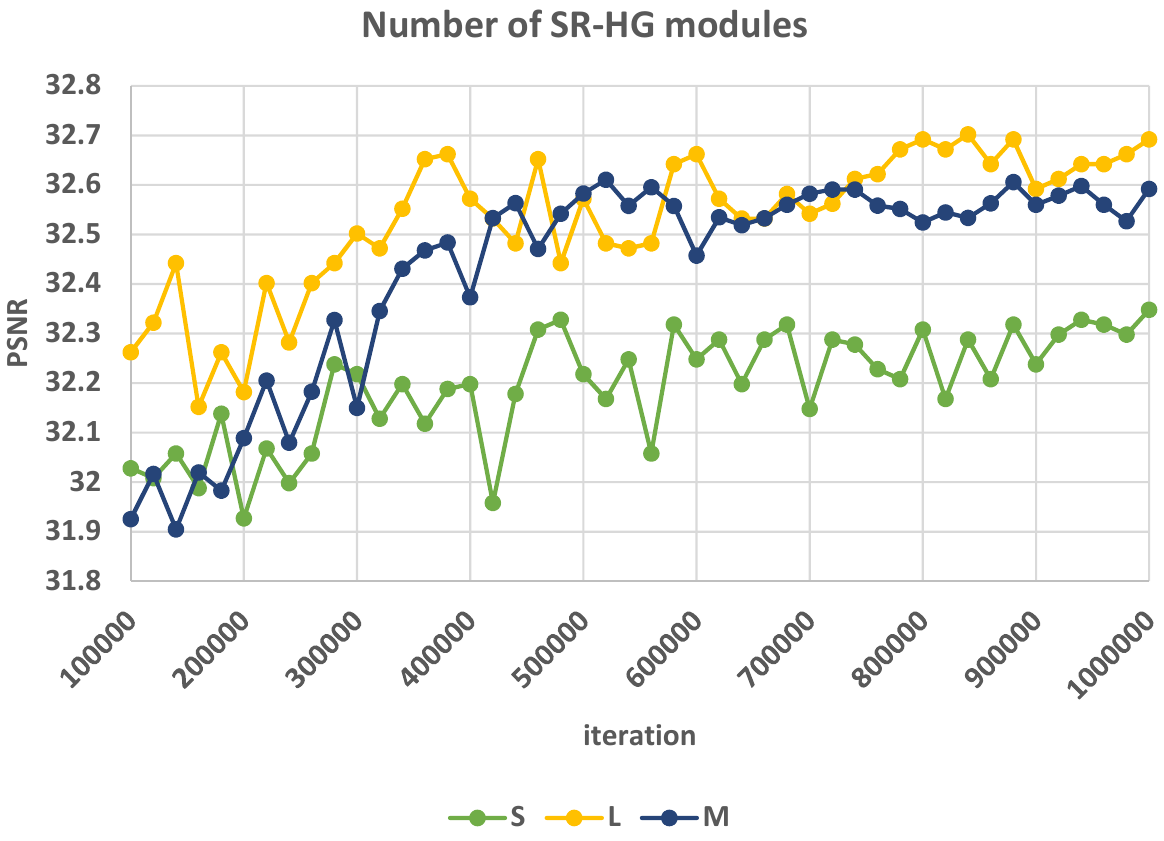}}
\caption{The number of SR-HG modules analysis. Note that we compare networks using different numbers of \textit{SR-HG} modules for repeated feature extraction.}
\label{Figure 7}
\end{center}
\vskip -0.3in
\end{figure}

From Figure~\ref{Figure 7}, we can see that network \textit{L} (4 SR-HG module) gives the highest PSNR result. For network \textit{S} (2 SR-HG module), its performance is lower than network \textit{M} and network \textit{L}. For network \textit{L} (4 SR-HG module), it requires extra 33\% parameters as compared with network \textit{M} but only achieves slight (0.1 dB) improvement in PSNR. This result shows that our proposed HBPN has the best trade-off between performance and number of parameters. To further study the significance of each SR-HG module, let us visualize the activation maps of the output of each SR-HG module in our HBPN network. 

\begin{figure}[htb]
\vskip 0.01in
\begin{center}
\centerline{\includegraphics[width=0.9\columnwidth]{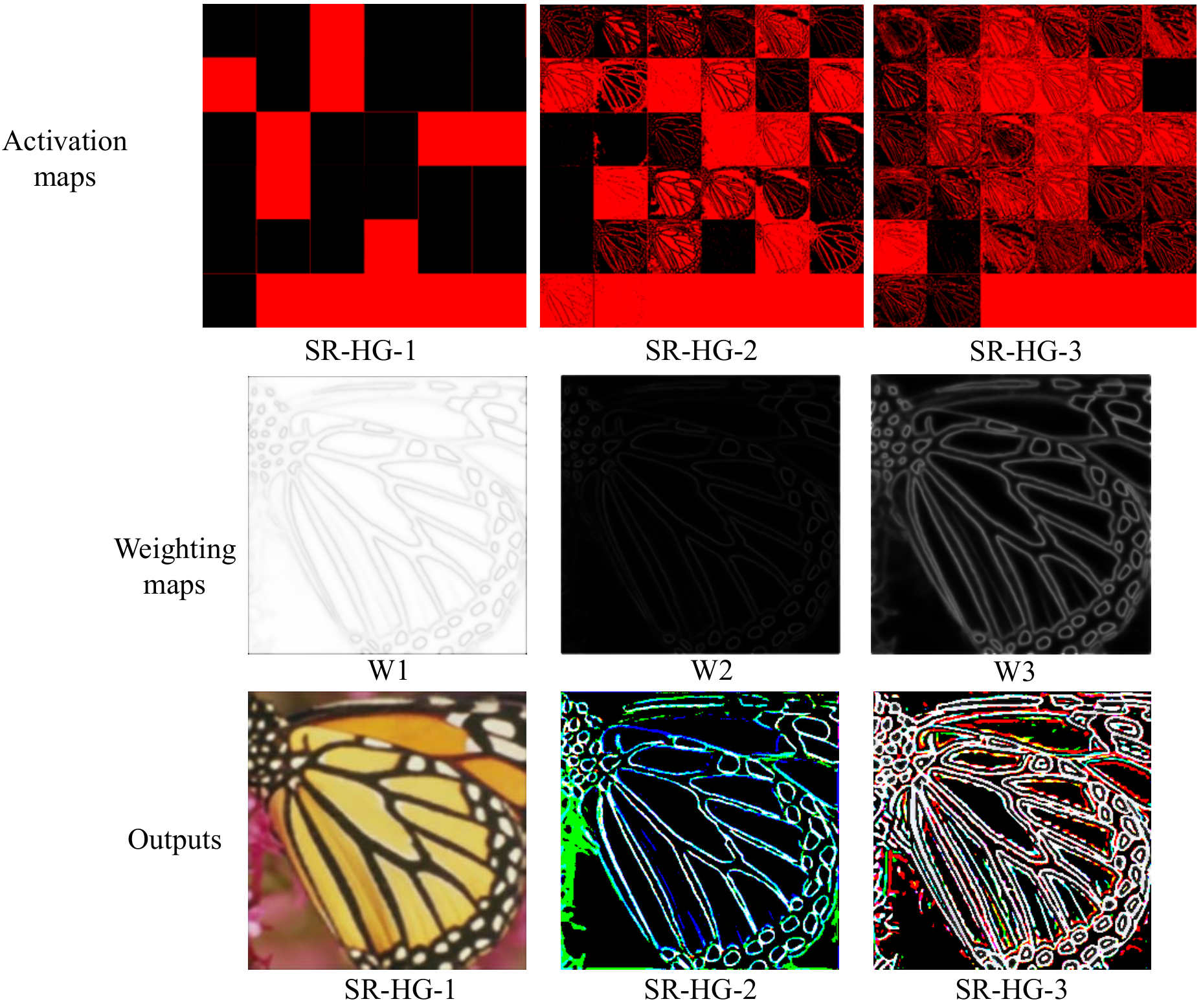}}
\caption{Visualization of activation maps, weighting maps and intermediate outputs. For better observation, please check the electronic version of this figure.}
\label{Figure 8}
\end{center}
\vskip -0.5in
\end{figure}

In the Figure~\ref{Figure 8}, the first row shows three activation maps of each SR-HG output on image \textit{butterfly}. We believe that the reason why CNNs outperform other patch-based learning approaches is that CNNs use activation layers to introduce the nonlinearity in the network to improve the feature representation power of filters. Hence, we show the activation maps rather than the output feature maps to show how activation layer works. In our design, we use the \textit{PReLU} function that assigns weight 1 to non-zero values and very small weights to negative values. We can visualize the weights as the activation maps. In our experiments, we chose the last \textit{PReLU} layer of each SR-HG module to make comparison. We can observe that the activation map of the \textit{SR-HG-1} module has high activation across some of the feature maps while zero activation on others because the first layer only focuses on reconstructing the low-frequency information on averaging the whole image. This can be observed on the output of \textit{SR-HG-1} of Figure~\ref{Figure 8}. For \textit{SR-HG-2} and \textit{SR-HG-3}, there are more activated values on the activation maps, that focus on edge and texture regions. We calculated the percentage of activated values on \textit{SR-HG-1}, \textit{SR-HG-2} and \textit{SR-HG-3} and found the value decreases from 30.55\%, 25.46\% to 22.35\%, which explains that the convolutional filters focus more on the edge and texture reconstruction.

\textbf{The effect of WR process.}
Finally, we compare the WR process and the plain concatenated process in Table~\ref{Table 1}. We design the plain concatenated process and WR process with the structure as shown in Figure~\ref{Figure 5}. They use the same SR-HG modules for feature extraction and the only difference is the final reconstruction process. The results were conducted on Set5, Set14 dataset of 2$\times$, 4$\times$ and 8$\times$ enlargement.

\begin{table}[t]
\caption{Comparison of the network using plain concatenation block or WR reconstruction block, including PSNR and SSIM for scale 2$\times$, 4$\times$ and 8$\times$ SR on Set5 and Set14. {\color{red}Red} indicates the best results.}
\label{Table 1}
\vskip -0.1in
\begin{center}
\begin{small}
\scalebox{0.8}{
\begin{tabular}{c|ccccc}
\hline
\multirow{2}{*}{Algorithm} & \multirow{2}{*}{Scale} & \multicolumn{2}{c}{Set5} & \multicolumn{2}{c}{Set14} \\
 &  & PSNR & SSIM & PSNR & SSIM \\ \hline
Plain model & 2 & 37.95 & 0.959 & 33.61 & 0.917 \\
WR model & 2 & {\color{red}38.13} & {\color{red}0.961} & {\color{red}33.78} & {\color{red}0.921} \\ \hline
Plain model & 4 & 32.33 & 0.889 & 28.55 & 0.731 \\
WR model & 4 & {\color{red}32.55} & {\color{red}0.900} & {\color{red}28.67} & {\color{red}0.785} \\ \hline
Plain model & 8 & 26.89 & 0.761 & 24.81 & 0.632 \\
WR model & 8 & {\color{red}27.17} & {\color{red}0.785} & {\color{red}24.96} & {\color{red}0.642} \\ \hline
\end{tabular}
}
\end{small}
\end{center}
\vskip -0.3in
\end{table}

\begin{table*}[t]
\caption{Quantitative evaluation of state-of-the-art SR approaches, including PSNR and SSIM for scale 2$\times$, 4$\times$ and 8$\times$. {\color{red}Red} indicates the best and {\color{blue}blue} indicates the second best results.}
\label{Table 2}
\vskip -0.1in
\begin{center}
\begin{small}
\scalebox{0.9}{
\begin{tabular}{lccccccccccc}
\hline
\multirow{2}{*}{Algorithm} & \multirow{2}{*}{Scale} & \multicolumn{2}{c}{Set5} & \multicolumn{2}{c}{Set14} & \multicolumn{2}{c}{BSD100} & \multicolumn{2}{c}{Urban100} & \multicolumn{2}{c}{Manga109} \\
 &  & PSNR & SSIM & PSNR & SSIM & PSNR & SSIM & PSNR & SSIM & PSNR & SSIM \\ \hline
Bicubic &  & 33.65 & 0.930 & 30.34 & 0.870 & 29.56 & 0.844 & 27.39 & 0.841 & 31.05 & 0.935 \\
A+~\cite{Timofte2015} &  & 36.54 & 0.954 & 32.40 & 0.906 & 31.22 & 0.887 & 29.23 & 0.894 & 35.33 & 0.967 \\
CRFSR~\cite{ICIP18} &  & 37.29 & 0.957 & 32.61 & 0.909 & 31.61 & 0.891 & 30.48 & 0.907 & 36.78 & 0.970 \\
SRCNN~\cite{SRCNN} &  & 36.65 & 0.954 & 32.29 & 0.903 & 31.36 & 0.888 & 29.52 & 0.895 & 35.72 & 0.968 \\
VDSR~\cite{VDSR} &  & 37.53 & 0.958 & 32.97 & 0.913 & 31.90 & 0.896 & 30.77 & 0.914 & 37.16 & 0.974 \\
DRRN~\cite{DRRN} & 2$\times$ & 37.74 & 0.959 & 33.23 & 0.913 & 32.05 & 0.897 & 31.23 & 0.919 & 37.92 & 0.976 \\
SRResNet~\cite{SRGAN} &  & - & - & - & - & - & - & - & - & - & - \\
LapSRN~\cite{LapSRN} &  & 37.52 & 0.959 & 33.08 & 0.913 & 31.80 & 0.895 & 30.41 & 0.910 & 37.27 & 0.974 \\
EDSR~\cite{EDSR} &  & {\color{blue}38.11} & {\color{blue}0.960} & {\color{red}33.92} & {\color{blue}0.919} & {\color{blue}32.32} & {\color{blue}0.901} & 32.93 & {\color{blue}0.935} & {\color{blue}39.10} & 0.977 \\
DBPN~\cite{DBPN} &  & 38.09 & {\color{blue}0.960} & {\color{blue}33.85} & {\color{blue}0.919} & 32.27 & 0.900 & {\color{blue}32.96} & 0.931 & {\color{blue}39.10} & {\color{blue}0.978} \\
HBPN(Ours) &  & {\color{red}38.13} & {\color{red}0.961} & 33.78 & {\color{red}0.921} & {\color{red}32.33} & {\color{red}0.902} & {\color{red}33.12} & {\color{red}0.938} & {\color{red}39.30} & {\color{red}0.979} \\ \hline
Bicubic &  & 28.42 & 0.810 & 26.10 & 0.704 & 25.96 & 0.669 & 23.64 & 0.659 & 25.15 & 0.789 \\
A+~\cite{Timofte2015} &  & 30.30 & 0.859 & 27.43 & 0.752 & 26.82 & 0.710 & 24.34 & 0.720 & 27.02 & 0.850 \\
CRFSR~\cite{ICIP18} &  & 31.10 & 0.871 & 27.87 & 0.765 & 27.05 & 0.719 & 24.89 & 0.744 & 28.12 & 0.872 \\
SRCNN~\cite{SRCNN} &  & 30.49 & 0.862 & 27.61 & 0.754 & 26.91 & 0.712 & 24.53 & 0.724 & 27.66 & 0.858 \\
VDSR~\cite{VDSR} &  & 31.35 & 0.882 & 28.03 & 0.770 & 27.29 & 0.726 & 25.18 & 0.753 & 28.82 & 0.886 \\
DRRN~\cite{DRRN} & 4$\times$ & 31.68 & 0.888 & 28.21 & 0.772 & 27.38 & 0.728 & 25.44 & 0.764 & 29.46 & 0.896 \\
SRResNet~\cite{SRGAN} &  & 32.05 & 0.891 & 28.53 & 0.780 & 27.57 & 0.735 & 26.07 & 0.784 & - & - \\
LapSRN~\cite{LapSRN} &  & 31.54 & 0.885 & 28.19 & 0.772 & 27.32 & 0.728 & 25.21 & 0.756 & 29.09 & 0.890 \\
EDSR~\cite{EDSR} &  & 32.46 & 0.897 & {\color{blue}28.80} & {\color{red}0.788} & 27.71 & {\color{blue}0.742} & {\color{blue}26.64} & {\color{blue}0.803} & 31.02 & {\color{blue}0.915} \\
DBPN~\cite{DBPN} &  & {\color{blue}32.47} & {\color{blue}0.898} & {\color{red}28.82} & {\color{blue}0.786} & {\color{blue}27.72} & 0.740 & 26.60 & 0.795 & {\color{blue}31.13} & 0.914 \\
HBPN(Ours) &  & {\color{red}32.55} & {\color{red}0.900} & 28.67 & 0.785 & {\color{red}27.77} & {\color{red}0.743} & {\color{red}27.30} & {\color{red}0.818} & {\color{red}31.57} & {\color{red}0.920} \\ \hline
Bicubic &  & 24.39 & 0.657 & 23.19 & 0.568 & 23.67 & 0.547 & 21.24 & 0.516 & 21.68 & 0.647 \\
A+~\cite{Timofte2015} &  & 25.52 & 0.692 & 23.98 & 0.597 & 24.20 & 0.568 & 21.37 & 0.545 & 22.39 & 0.680 \\
CRFSR~\cite{ICIP18} &  & 26.07 & 0.732 & 23.97 & 0.600 & 24.20 & 0.569 & 21.36 & 0.550 & 22.59 & 0.688 \\
SRCNN~\cite{SRCNN} &  & 25.33 & 0.689 & 23.85 & 0.593 & 24.13 & 0.565 & 21.29 & 0.543 & 22.37 & 0.682 \\
VDSR~\cite{VDSR} &  & 25.72 & 0.711 & 24.21 & 0.609 & 24.37 & 0.576 & 21.54 & 0.560 & 22.83 & 0.707 \\
DRRN~\cite{DRRN} & 8$\times$ & 26.18 & 0.738 & 24.42 & 0.622 & 24.59 & 0.587 & 21.88 & 0.583 & 23.60 & 0.742 \\
SRResNet~\cite{SRGAN} &  & - & - & - & - & - & - & - & - & - & - \\
LapSRN~\cite{LapSRN} &  & 26.15 & 0.738 & 24.35 & 0.620 & 24.54 & 0.586 & 21.81 & 0.582 & 23.39 & 0.735 \\
EDSR~\cite{EDSR} &  & 26.97 & 0.775 & 24.94 & 0.640 & 24.80 & 0.596 & 22.47 & 0.620 & 24.58 & 0.778 \\
DBPN~\cite{DBPN} &  & {\color{red}27.21} & {\color{blue}0.784} & {\color{red}25.13} & {\color{red}0.648} & {\color{blue}24.88} & {\color{blue}0.601} & {\color{blue}22.69} & {\color{blue}0.622} & {\color{blue}24.96} & {\color{blue}0.799} \\
HBPN(Ours) &  & {\color{blue}27.17} & {\color{red}0.785} & {\color{blue}24.96} & {\color{blue}0.642} & {\color{red}24.93} & {\color{red}0.602} & {\color{red}23.04} & {\color{red}0.647} & {\color{red}25.24} & {\color{red}0.802} \\ \hline
\multicolumn{12}{c}{NTIRE2019 Validation} \\
\hline
\multicolumn{4}{c}{Algorithm} & \multicolumn{4}{c}{PSNR} & \multicolumn{4}{c}{SSIM} \\
\hline
\multicolumn{4}{c}{Bicubic} & \multicolumn{4}{c}{29.548} & \multicolumn{4}{c}{0.844} \\
\multicolumn{4}{c}{Using the proposed plain HBPN} & \multicolumn{4}{c}{{\color{blue}33.41}} & \multicolumn{4}{c}{{\color{blue}0.889}} \\
\multicolumn{4}{c}{HBPN with Weighted Reconstruction} & \multicolumn{4}{c}{{\color{red}33.88}} & \multicolumn{4}{c}{{\color{red}0.920}} \\ \hline
\end{tabular}
}
\end{small}
\end{center}
\vskip -0.3in
\end{table*}

From Table~\ref{Table 1}, we can see that using WR process can significantly improve the PSNR by at least 0.11 dB. The effectiveness of WR process can be further explained in Figure~\ref{Figure 8}. In the second and third rows of Figure~\ref{Figure 8}, we visualize the weighting maps of each SR-HG module and coarse SR outputs. For the first SR-HG module, the weighting map focuses on the low-frequency domain that reconstructs the main components of the image. For the second and third SR-HG modules, the weighting maps give high attentions to the edge regions. From the coarse SR output of each SR-HG module, we can also match the results with their weighting maps. Note that the output of \textit{SR-HG-2} focuses on the edge reconstruction on G and B channels and the output of \textit{SR-HG-3} focuses on the edge reconstruction on the R channel. From the aspect of gradient based edge detection, \textit{SR-HG-2} focuses on the first-order edge reconstruction (see the single-line edges on the output of \textit{SR-HG-2}) while \textit{SR-HG-3} pays attention on the second-order edge reconstruction (see the double-line edges on the output of \textit{SR-HG-3}). This can prove that using more SR-HG modules can explore deeper features in terms of the order of the pixel gradient.  

From Table~\ref{Table 1}, it can be found that using WR process is very efficient that can gain 0.2 dB and 0.1 higher than the plain process in terms of PSNR and SSIM, respectively. We also show the weighting maps to possibly indicate the contribution of coarse SR results. We name the weighting maps at different stages of SR-HG modules as \textit{W}1, \textit{W}2 and \textit{W}3. The weighting maps are visualized by normalizing the pixel values in the range of [0, 255]. The weighting map corresponds to the SR-HG results giving different weights to the pixel values. The first weighting map gives a large weights on the whole image and small weights on the edges. The second and third weighting maps give higher weights to the non-edge regions (first-order edge detection) and edge regions (second-order edge detection), respectively.
\begin{figure*}[t]
\vskip 0.01in
\begin{center}
\centerline{\includegraphics[width=0.7\textwidth]{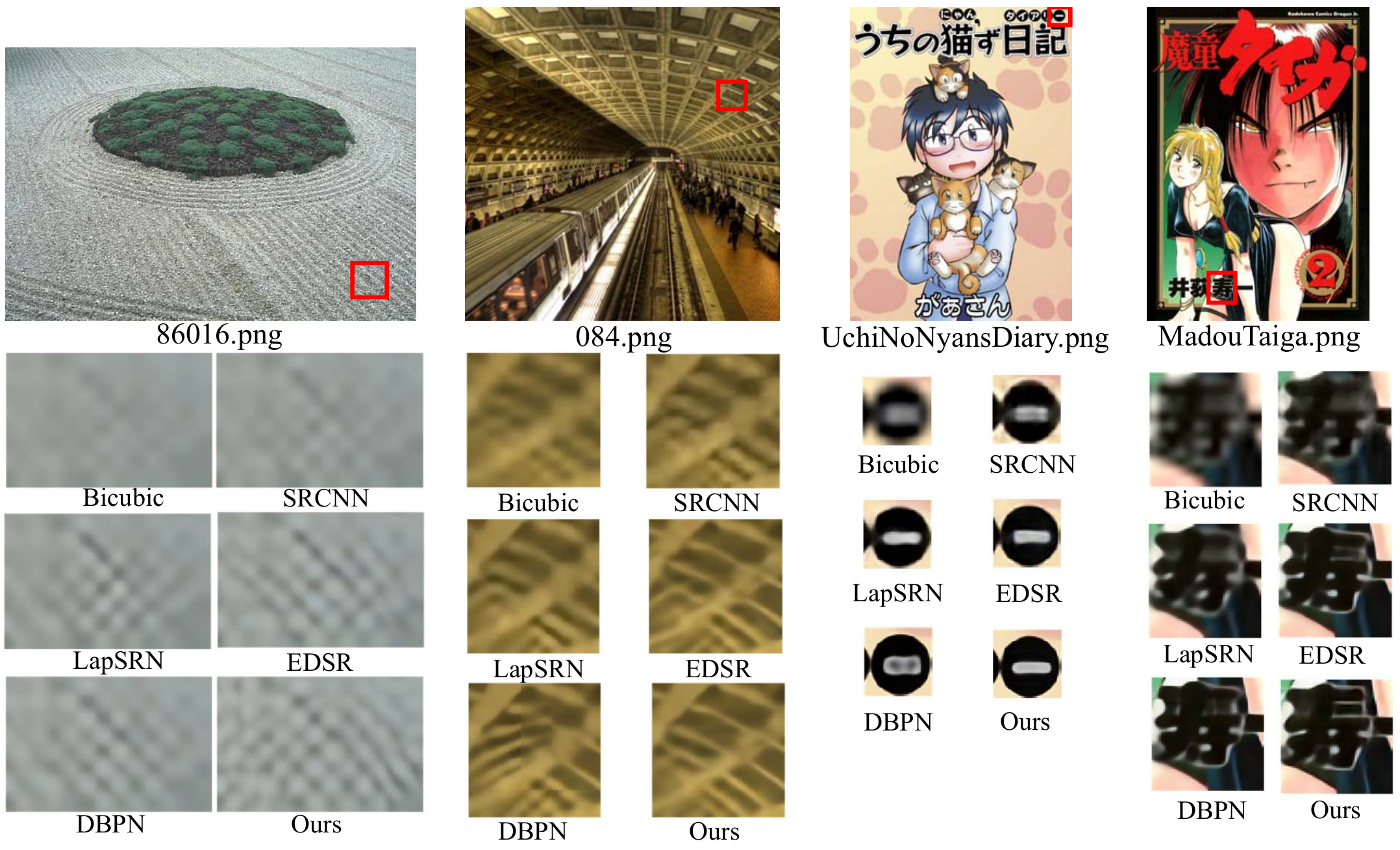}}
\caption{Visual quality comparison among different SR algorithms on 8$\times$ super-resolution.}
\label{Figure 9}
\end{center}
\vskip -0.3in
\end{figure*}
\subsection{Comparison with the state-of-the-art SR approaches}
To prove the effectiveness of the proposed methods, we conducted experiments by comparing with most (if not all) state-of-the-art SR algorithms: Bicubic, A+ ~\cite{Timofte2015}, CRFSR ~\cite{ICIP18}, SRCNN ~\cite{SRCNN}, VDSR ~\cite{VDSR}, DRCN ~\cite{DRCN}, LapSRN ~\cite{LapSRN}, SRResNet ~\cite{SRGAN}, EDSR ~\cite{EDSR} and DBPN ~\cite{DBPN}. PSNR and SSIM are used to evaluate the proposed method and others. Generally, PSNR and SSIM are calculated by converting RGB image to YUV and only the Y-channel image taken for consideration. During the testing, we rotated and flipped LR images for augmentation to generate several augmented inputs, and then applied the inverse transform and averaged all the outputs together to form the final SR results. For different scaling factors $\mathit{s}$, we exclude $\mathit{s}$ pixels at boundaries to avoid boundary effect. For SR results, SRCNN, VDSR, SRResNet, EDSR and DBPN were reimplemented and provided by the authors of~\cite{DBPN} and LapSRN was provided by the authors of~\cite{LapSRN}. Note that, this of our proposed approach also participated in the NTIRE2019 Real Image Super-resolution Challenge~\cite{NTIRE2019}. Table~\ref{Table 2} also includes the validation testing results of this dataset. For this competition, it targets at real daily images, with down-sampling process using different degradation and distortions, and all images were taken by DSLR cameras in natural environments. However, all the state-of-the-art SR algorithms in the literature have been trained by using bicubic down-sampled images. It would then be inappropriate to use our HBPN model to make comparison with approaches in the literature with the NTIRE2019 validation dataset. Hence we just mainly listed out the results of our model using or without using the final stage of the proposed Weighted Reconstruction model for comparison. For more visual quality comparison, it is available at \url{https://github.com/Holmes-Alan/HBPN}.

We show the quantitative results in Table~\ref{Table 2}. Our proposed HBPN method outperforms other state-of-the-art approaches in all scales. Among these approaches, our proposed work can outperform EDSR and DBPN by large improvement (0.1-0.6 dB) on $8\times$ enlargement and improve the SR quality about 0.1-0.4 dB on $2\times$ and $4\times$ enlargement. Note that the PSNR and SSIM on BSD100, Urban100 and Manga109 using DBPN are different from~\cite{DBPN} because we calculated the results on the whole image (rather than dividing images into four parts and calculating separately) by running their released code for fair comparison. For visual comparison, $2\times$ and $4\times$ enlargement are difficult to distinguish the improvement of the proposed method. We show $8\times$ enlargement in Figure~\ref{Figure 9}, including the \textit{86016.png} image from BSD100, \textit{084.png} image from Urban100 and \textit{UchiNoNyansDiary.png and MadouTaiga.png} images from Manga109. Figure~\ref{Figure 9} shows that both DBPN and EDSR cannot reconstruct well the fine texture of \textit{86016.png}. On the other hand, our result can predict a clearer pattern of the sand. On the edge pattern of the roof on \textit{084.png}, DBPN fails to reconstruct the concrete texture. Our approach can predict the horizontal and diagonal strides of the roof. The last two images of our approach on Manga109 give better visual quality in comparison with different approaches. On \textit{UchiNoNyansDiary.png}, there is a Japanese character on the right upper corner that cannot be clearly reconstructed by SRCNN and LapSRN. DBPN, on the other hand, gives a result containing holes on that stride that is misunderstanding. Our result actually can predict a sharper character. Similarly on \textit{MadouTaiga.png}, the Japanese character inside the red box can better be observed on our result. Other SR approaches either generate blur edges on the strides or miss the stride pattern.

From all the results, we can see that our proposed HBPN approach can achieve better SR performance both quantitatively and qualitatively. It not only preserves the edge components, but also reconstructs the fine textures at different scaling factors. 
\section{Conclusion}

We have proposed a Hierarchical Back Projection Network for image Super-Resolution on different up-scaling factors. Different from the previous SR study, we focus on feature extraction by conducting a HourGlass structure to learn the features in a bottom-up and top-down manner. The back projection mechanism is embedded into the network to update the low-resolution and high-resolution feature maps to reduce the errors. Meanwhile, we propose a self-weighting process that each HourGlass module generates one intermediate SR result along with its weighting map. By using the proposed Weighted Reconstruction block, we normalize the weighting maps to tune the contribution of each intermediate SR results for generating the final SR images. Results on quantitative and quality evaluation show its advantages over other approaches. Furthermore, we have also visualized the trained feature maps to illustrate the power of feature representation of each HourGlass module.

{\small
\bibliographystyle{ieee_fullname}
\bibliography{paper_draft_bib}
}

\end{document}